\documentclass[conference]{IEEEtran}

\usepackage{amsmath,amssymb}
\usepackage{booktabs}
\usepackage{float}
\usepackage{graphicx}
\usepackage{microtype}
\usepackage{xcolor}
\usepackage{url}
\usepackage[colorlinks=true,allcolors={blue!65!black},breaklinks=true]{hyperref}

\graphicspath{{figures/}}
\newcommand{\Dforget}{D_{\mathrm{f}}}
\newcommand{\Dretain}{D_{\mathrm{r}}}
\newcommand{\Dunmatched}{D_{\mathrm{u}}}
\newcommand{\Dmatched}{D_{\mathrm{m}}}
\newcommand{\Lloss}{\mathcal{L}}
\newcommand{\thetaft}{\theta_{\mathrm{ft}}}
\newcommand{\thetaun}{\theta_{\mathrm{un}}}
\newcommand{\thetart}{\theta_{\mathrm{rt}}}
\newcommand{\thetanr}{\theta_{\mathrm{nr}}}
\newcommand{\tsf}{\mathrm{TSF}}

\title{Deletion You Can Check:\\Auditing LLM Unlearning}
\author{\IEEEauthorblockN{Mayank Gulati and Gerhard Wunder}
  \IEEEauthorblockA{Freie Universit\"at Berlin\\
  \{mayank.gulati, gerhard.wunder\}@fu-berlin.de}}

\begin{document}
\pagestyle{plain}
\maketitle

\begin{abstract}
A common indication that a fine-tuned model has forgotten a deleted record is
a membership-inference score near chance (ROC-AUC $\approx 0.5$). This reading
is sound only if the comparison population represents how the record should
behave after a valid deletion---a counterfactual that is rarely validated. We
test this premise in a controlled $2\times2$ study pairing two
membership-inference benchmarks, ArxivMIA and BookMIA, with two language
models, Llama-3.2-1B and Qwen2.5-1.5B. Exact retraining validates the
membership reference: it lies within the
prespecified chance band on ArxivMIA but reaches $0.122$ on BookMIA, causing
the same metric to reject an empirical deletion reference. A post-hoc
BookMIA/Qwen diagnostic replaces benchmark label-0 records with same-pool,
never-fine-tuned records and moves the reference from $0.257$ to $0.523$,
supporting population mismatch in that cell. Each approximate-unlearning
checkpoint is judged jointly on membership removal, deletion-targeted shift
beyond non-deletion variation, and utility. On the
admitted ArxivMIA/Llama axis, no tested configuration of gradient ascent with
retain descent (GA+GD), negative-preference optimization with retain training
(NPO-RT), or teacher--student unlearning (SCRUB) passes. A score near chance is
not evidence of forgetting until a reference establishes what chance means.
\end{abstract}


\section{Introduction}
\label{sec:introduction}

Organizations increasingly adapt pretrained large language models (LLMs) to
proprietary or user-contributed corpora through
fine-tuning~\cite{hu2022lora,dettmers2023qlora}. A later deletion request
creates a gap between removing data and verifying its effect on a trained
model. Deleting the source record prevents its future use, but does not
establish whether an existing fine-tuned checkpoint still reflects that
record. Article~17 of the General Data Protection Regulation (GDPR), for
example, motivates erasure in specified circumstances but does not define how
to test a trained model for residual influence~\cite{gdpr2016}. We therefore
ask: after the model is updated, what evidence shows that it behaves as if the
requested records had not been used for fine-tuning?

We answer this question in the setting of \emph{controlled fine-tuning
deletion}. Here, the evaluator knows which records entered adaptation and can
repeat the procedure without the forget set. This observability is unavailable
for an unknown pretraining corpus. The broader principle---validate a metric
against a counterfactual with known deletion status---may extend beyond
fine-tuning, but our evidence and thresholds do not.

Within this setting, exact retraining provides an empirical deletion reference:
the same fine-tuning procedure starts from the same base checkpoint but omits
the forget set. We do not present retraining as a formal certificate or require
it for every deletion request. If it is affordable for the request at hand,
the retrained model can itself be deployed; approximate methods matter because
repeating that remedy is often costly~\cite{cao2015towards,
bourtoule2021machine}. Our use is narrower: a bounded set of retraining runs
serves as an evaluation control for the metric used to judge cheaper
approximate methods~\cite{kurmanji2023towards,zhang2024negative}. The cost is
incurred when validating the audit, not for every deletion request. If an
evaluator cannot run even a representative set of these controls, the audit
cannot provide the same level of evidence.

This control is necessary because a common evaluation contains a hidden
assumption. It compares the requested records with ``unseen'' records and
interprets membership ROC-AUC near $0.5$ as
forgetting~\cite{shokri2017membership,carlini2022membership}. The conclusion is valid
only if the unseen population represents how the requested records should
behave after deletion. Otherwise, an AUC far from chance need not imply failed
deletion, and an AUC near chance need not imply successful deletion.

We observe exactly this failure on BookMIA/Qwen. Exact retraining omits the
forget set, yet the membership metric assigns it an AUC of $0.257$ and rejects
the empirical reference. The comparison mixes two notions: BookMIA labels
likely base-model \emph{pretraining} membership, whereas our deletion claim
concerns added \emph{fine-tuning} membership. In a post-hoc diagnostic,
replacing benchmark label-0 records with same-pool, never-fine-tuned records
moves the reference to $0.523$. This one-cell result supports population
mismatch in the tested BookMIA/Qwen setting, but does not show that BookMIA
generally causes the reference failure.

\begin{figure*}[t]
\centering
\includegraphics[width=0.98\textwidth]{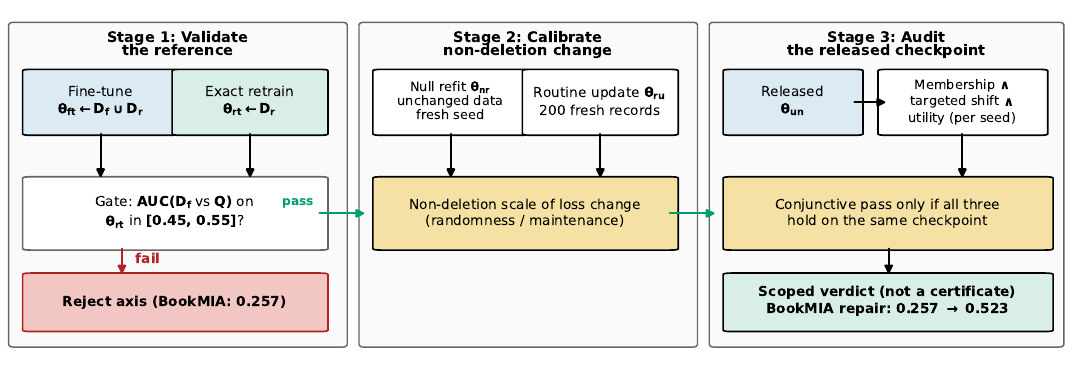}
\caption{\textbf{The reference-gated audit.} Stage 1 admits a membership axis
only if exact retraining reads chance on the proposed comparison population;
Stage 2 calibrates loss change that occurs without deletion; Stage 3
evaluates membership, targeted shift, and utility conjunctively per
checkpoint. The BookMIA/Qwen values are the post-hoc comparison-population
diagnostic in Figure~\ref{fig:reference}b, illustrating axis rejection and
repair rather than method acceptance.}
\label{fig:pipeline}
\end{figure*}

This failure leads to our \emph{reference-gated audit}
(Figure~\ref{fig:pipeline}). The audit follows a fixed sequence: validate the
membership comparison with exact retraining; calibrate non-deletion movement
using independent refits and routine updates; judge membership removal,
targeted shift, and utility jointly within each checkpoint; and finally
interpret detector performance under matched access and deployment
prevalence.

This sequence yields four contributions:
\begin{itemize}
    \item \textbf{We identify a benchmark reference failure.} In a
    $2\times2$ exact-retraining study, membership AUC ranges from $0.122$ to
    $0.474$; a post-hoc same-pool diagnostic moves one rejected axis from
    $0.257$ to $0.523$, supporting comparison-population mismatch in the
    tested BookMIA/Qwen cell.
    \item \textbf{We define a reference-gated audit for LLM fine-tuning
    unlearning.} The audit admits the membership axis only on a validated
    counterfactual, calibrates non-deletion movement with null refits and
    routine updates, and applies membership, targeted shift, and utility
    criteria conjunctively within each seed.
    \item \textbf{We audit representative approximate methods.} On the
    admitted ArxivMIA/Llama axis, no tested GA+GD, NPO-RT, or SCRUB
    configuration passes the conjunction; the original BookMIA/Qwen axis
    cannot support a method verdict, and its post-hoc repair does not rescue
    the tested aggressive GA+GD setting. A membership-only criterion accepts
    checkpoints with severely degraded retained utility.
    \item \textbf{We characterize operational auditability.} Matched-access
    comparisons and 1:100-prevalence projections show that detector rankings
    are unstable across settings and that balanced AUC does not determine
    operational precision.
\end{itemize}

These contributions define an empirical assurance procedure, not a formal
unlearning guarantee and not a claim that every approximate method fails.
The actionable conclusion: before judging an approximate unlearning method,
an evaluator should first show that the evaluation axis behaves correctly on
controls with known ground truth.\\

\textbf{Public source code and artifact:}
It includes the configurations, per-example scores, report and figure scripts, environment
specification, and code used to produce the reported results.
\url{https://github.com/MakGulati/deletion-you-can-check}.

\section{Related Work}
\label{sec:related}

\textbf{Approximate unlearning for LLMs.} Approximate methods replace full
retraining with cheaper targeted updates~\cite{cao2015towards,
bourtoule2021machine}. For language models, SCRUB trains a student to preserve
teacher behavior on retained data while diverging on the forget
set~\cite{kurmanji2023towards}; NPO instead uses a bounded preference objective
over negative examples~\cite{zhang2024negative}. These methods ask how to
produce a useful unlearned model. We ask the complementary question: what
evidence makes a claim about that model credible? Accordingly, these
methods serve as test cases for our audit; we do not propose a new unlearning
optimizer.

\textbf{Retraining-relative and multidimensional evaluation.} Exact retraining
is commonly used as a behavioral reference, although behavioral similarity
alone cannot establish dataset history~\cite{thudi2022necessity}. TOFU and
MUSE broaden LLM-unlearning evaluation across forgetting and utility
dimensions~\cite{maini2024tofu,shi2025muse}. Broader coverage, however, does
not establish that each metric represents the intended deletion
counterfactual. Our audit therefore validates the membership axis on exact
retraining before combining membership, targeted shift, and utility within a
checkpoint; null refits and routine updates separately calibrate movement that
occurs without deletion.

\textbf{Membership inference and comparison populations.} Membership attacks
distinguish training from nontraining behavior~\cite{shokri2017membership,
carlini2022membership}; difficulty-calibrated attacks likewise depend on
matched reference distributions~\cite{watson2022difficulty}. BookMIA
illustrates the difficulty of constructing credible pretraining-membership
ground truth for language models~\cite{shi2024detecting}. In our controlled
fine-tuning setting, however, a benchmark ``non-member'' may encode base-model
pretraining status rather than omission from the added fine-tuning set. We
separate these populations and admit a membership axis only when exact
retraining validates the proposed comparison.

\textbf{Deletion-history inference and audit assumptions.} FUMA identifies
forgotten records from post-unlearning loss, gradient, and text
signals~\cite{deepak2025fuma}. RULI uses per-sample likelihood tests to
distinguish privacy leakage from retraining-relative efficacy and exposes the
limits of average-case and accuracy-only evaluations~\cite{naderloui2025rectifying}.
A SoK compares dataset audits by model access and required capabilities,
including dataset modification and auxiliary data~\cite{du2025sok}. Together,
these studies show that deletion history can remain detectable and conclusions
depend on access. We therefore separate current membership from deletion
history, match detector access, and report low-prevalence precision. Our
two-checkpoint procedure is an authorized audit, not a black-box attack;
balanced AUC does not imply operational precision.

\section{Setting and Access Model}
\label{sec:setting}

\subsection{Controlled fine-tuning deletion}

\textbf{Benchmark labels.} ArxivMIA and BookMIA label records as likely
members (1) or nonmembers (0) of the base model's pretraining
data~\cite{shi2024detecting,zhliu2024arxivmia}. These benchmark labels are
distinct from membership in our controlled fine-tuning set.

\textbf{Training and deletion reference.} Let $\theta_0$ be the pretrained
base model. For each seed $s$, we draw disjoint forget and retain sets,
$\Dforget$ and $\Dretain$, from the label-1 pool. We denote the fixed QLoRA
fine-tuning procedure by $\mathcal A(\theta_0,D;\xi)$, where $\xi$ denotes
training randomness. QLoRA freezes the quantized base model and trains
low-rank adapters~\cite{hu2022lora,dettmers2023qlora}. We construct
\begin{equation}
 \theta_{\mathrm{ft}}^{(s)}
 =\mathcal A(\theta_0,\Dforget\cup\Dretain;\xi_s),\qquad
 \theta_{\mathrm{rt}}^{(s)}
 =\mathcal A(\theta_0,\Dretain;\xi_s).
 \label{eq:training-references}
\end{equation}
For a request targeting $\Dforget$, an approximate method maps
$\thetaft^{(s)}$ to $\thetaun^{(s)}$. Matching the base checkpoint,
quantization, adapter, optimizer, epoch count, and seed makes
$\thetart^{(s)}$ the empirical deletion reference: it excludes $\Dforget$
from controlled fine-tuning but says nothing about unknown pretraining.

\textbf{Comparison populations.} The \emph{unmatched} set $\Dunmatched$
contains label-0 records, unlike the label-1 forget and retain sets. The
\emph{matched} set $\Dmatched$ contains disjoint, unused label-1 records from
their source pool, matching both benchmark label and source. Both sets are
excluded from fine-tuning; $\Dmatched$ appears only in the post-hoc BookMIA/Qwen
diagnostic.

\textbf{Loss and membership score.} For record $z$, let $z_t$ be its $t$th
scored token, $c_{z,t}$ its preceding context, and $T_z$ its scored length.
We use the standard mean token negative log-likelihood:
\begin{align}
 \Lloss(\theta;z)
 &=-\frac{1}{T_z}\sum_{t=1}^{T_z}\log p_\theta(z_t\mid c_{z,t}),\nonumber\\
 \Lloss(\theta;P)
 &=\frac{\sum_{z\in P}T_z\Lloss(\theta;z)}{\sum_{z\in P}T_z}.
 \label{eq:population-loss}
\end{align}
Thus, $\Lloss(\theta;P)$ is the mean over all scored tokens. We use the
base-calibrated membership score
\begin{equation}
 s_{\mathrm{mem}}(z;\theta)=
 \frac{\Lloss(\theta;z)}{\max\{\Lloss(\theta_0;z),\epsilon\}},
 \label{eq:membership-score}
\end{equation}
where $\epsilon>0$ prevents division by zero. This ratio calibrates record
difficulty against the base model~\cite{watson2022difficulty}, but not
population mismatch. For finite $P$, $s_{\mathrm{mem}}(P;\theta)$ is the
vector of per-record scores.

\subsection{What a verdict must cover}

An unlearning verdict must keep six questions separate:
\begin{itemize}
 \setlength{\itemsep}{0pt}
 \setlength{\parsep}{0pt}
 \setlength{\topsep}{2pt}
 \item \emph{Current membership:} Does the released model distinguish a record
 from a valid never-fine-tuned counterfactual?
 \item \emph{Deletion history:} Does the pre-to-post checkpoint change reveal
 which known members were targeted for deletion?
 \item \emph{Targeted shift:} Did forget records move toward exact retraining
 without comparable retained-record movement?
 \item \emph{Model utility:} Are retained and held-out losses preserved?
 \item \emph{Detector capability:} At fixed access and operating point, how well
 can a detector rank or recover targets?
 \item \emph{Detector access:} Which checkpoints, outputs, gradients, and
 population information does the detector require?
\end{itemize}
These properties are not interchangeable: global damage can erase membership
without selective forgetting, while two-checkpoint detection can reveal
deletion history after current membership disappears. Interpretation depends
on authorization: an owner or regulator may hold both checkpoints; an ordinary
API user does not.

\subsection{Access bands}

The access bands are nested: $\mathcal O_{\mathrm{rel}}\subseteq
\mathcal O_{\mathrm{2cp}}\subseteq\mathcal O_{\mathrm{wb}}$. A
\emph{released-model evaluator} observes only $\thetaun$ outputs or losses; an
\emph{authorized two-checkpoint auditor} also holds $\thetaft$ and measures
per-record changes; and a \emph{white-box auditor} additionally observes
parameters, gradients, and checkpoint displacement. We compare detectors only
under matched access; these audit capabilities do not imply adversarial access
to both checkpoints.

\section{Reference-Gated Audit}
\label{sec:audit}

The audit follows the order in which its evidence should be interpreted.
First, a reference gate tests whether exact retraining validates the chosen
membership score and comparison population. Second, calibration controls
measure changes that can occur without deletion. Third, on an admitted
membership axis, each candidate checkpoint is tested for membership removal,
targeted shift, utility preservation, and numerical validity. All four checks
must pass for the same seed and checkpoint. We assess detector access and
low-prevalence precision separately from this checkpoint verdict.

\subsection{Reference gate}

Let $\mathcal S$ be the evaluated seed set and let $Q$ be the comparison
population. We use $\widehat{\mathrm{AUC}}(x,y)$ for the empirical ROC-AUC
that distinguishes the records represented by score vector $x$ from those
represented by $y$. After exact retraining, a valid deletion counterfactual
should be indistinguishable from $\Dforget$ under the chosen score. We
therefore average the reference AUC across seeds:
\begin{equation}
 a_{\mathrm{rt}}(Q)=\frac{1}{|\mathcal S|}\sum_{s\in\mathcal S}
 \widehat{\mathrm{AUC}}\!\left(
 s_{\mathrm{mem}}(\Dforget;\thetart^{(s)}),
 s_{\mathrm{mem}}(Q;\thetart^{(s)})\right).
 \label{eq:reference-auc}
\end{equation}
We admit the score--population axis only when this average lies in the fixed
chance band:
\begin{equation}
 C_{\mathrm{mem}}(Q)=
 \mathbf 1\!\left\{0.45\le a_{\mathrm{rt}}(Q)\le0.55\right\}=1.
 \label{eq:reference-gate}
\end{equation}
The band is a preregistered engineering tolerance, not an inferential
equivalence interval. If Eq.~\eqref{eq:reference-gate} fails, the comparison
does not provide a valid membership-removal test, so we reject that axis before
judging any method. Passing only makes the axis admissible: global damage can
also erase membership discrimination, which is why targeted shift and utility
remain separate requirements.

\subsection{Calibration controls}

Exact retraining supplies the reference gate; three additional checkpoints
show how much movement can arise without approximate unlearning. Let
$\widetilde\xi_s$ be training randomness independent of $\xi_s$, let
$D_{\mathrm{new}}$ contain 200 fresh records, and let
$\mathcal M(\theta,D;\zeta)$ be a routine-update procedure with randomness
$\zeta$. We construct
\begin{align}
 \thetanr^{(s)}
 &=\mathcal A(\theta_0,\Dforget\cup\Dretain;\widetilde\xi_s),\nonumber\\
 \theta_{\mathrm{ru}}^{(s)}
 &=\mathcal M(\thetaft^{(s)},D_{\mathrm{new}};\zeta_s),\nonumber\\
 \theta_{\mathrm{rt+ru}}^{(s)}
 &=\mathcal M(\thetart^{(s)},D_{\mathrm{new}};\zeta_s).
 \label{eq:audit-controls}
\end{align}
Each checkpoint answers a different null question. The \emph{null refit}
$\thetanr^{(s)}$ repeats the original fine-tuning data with independent
randomness and measures refit variation. The \emph{routine update}
$\theta_{\mathrm{ru}}^{(s)}$ measures maintenance-induced change when no
record is deleted. Applying the same update after exact retraining yields
$\theta_{\mathrm{rt+ru}}^{(s)}$ and tests whether maintenance moves the
deletion reference itself. These controls calibrate the interpretation of
observed movement; they do not alter the candidate thresholds below.

\subsection{Membership-removal criterion}

For candidate checkpoint $\thetaun^{(s)}$, we measure how well the fixed
membership score separates $\Dforget$ from $Q$:
\begin{align}
 A_s(Q)
 &=\widehat{\mathrm{AUC}}\!\left(
 s_{\mathrm{mem}}(\Dforget;\thetaun^{(s)}),\right.\nonumber\\[-2pt]
 &\hspace{39mm}\left.s_{\mathrm{mem}}(Q;\thetaun^{(s)})\right).
 \label{eq:membership-auc}
\end{align}
If $C_{\mathrm{mem}}(Q)=0$, we may report $A_s(Q)$ diagnostically, but we do
not use it to judge the method. On an admitted axis,
\begin{equation}
 \mathsf{Mem}_s=\mathbf 1\{0.45\le A_s(Q)\le0.55\}.
 \label{eq:membership-pass}
\end{equation}
An AUC near $0.5$ provides little ranking separation between forget and
comparison records under this fixed score. Values near either $0$ or $1$
indicate strong separation, so the acceptance interval is necessarily
two-sided. As with the reference gate, this interval was fixed after
exploratory work and before confirmatory runs; it is an engineering tolerance,
not a statistical equivalence test.

\subsection{Targeted-shift criterion}

A candidate can erase membership discrimination by damaging the model rather
than selectively removing the requested records. We therefore compare its
movement with exact retraining on both forget and retain sets. The
\emph{targeted shift fraction} (TSF) expresses the candidate's loss change on
population $P$ in units of the exact-retraining change on $\Dforget$:
\begin{align}
 \Delta_{f,s}^{\mathrm{rt}}
 &=\Lloss(\thetart^{(s)};\Dforget)
   -\Lloss(\thetaft^{(s)};\Dforget),\\
 \tsf_s(P)
 &=\frac{\Lloss(\thetaun^{(s)};P)
   -\Lloss(\thetaft^{(s)};P)}{\Delta_{f,s}^{\mathrm{rt}}},
 \quad P\in\{\Dforget,\Dretain\}.
 \label{eq:tsf}
\end{align}
On $\Dforget$, TSF $=0$ means no movement from the fine-tuned checkpoint,
whereas TSF $=1$ reproduces the direction and magnitude of the
exact-retraining loss change. On $\Dretain$, a value near zero means little
collateral movement on the same scale. We define
\begin{equation}
 \mathsf{Sel}_s=\mathbf 1\!\left\{\tsf_s(\Dforget)\ge0.50
 \ \land\ |\tsf_s(\Dretain)|\le0.20\right\}.
 \label{eq:selectivity-predicate}
\end{equation}
Thus, the forget set must achieve at least half of the exact-retraining
movement while absolute retain-set movement remains within $20\%$ of that
common scale. We evaluate TSF only when
$|\Delta_{f,s}^{\mathrm{rt}}|\ge10^{-3}$; otherwise the denominator is too
small for reliable normalization and $\mathsf{Sel}_s$ is false. We use the
forget-set denominator for both populations because a retain-specific
exact-retraining change can approach zero.

\subsection{Utility-preservation criterion}

Targeted movement is insufficient if the released model loses utility. For
$P\in\{\Dretain,D_{\mathrm{test}}\}$, we therefore compare candidate
perplexity with the fine-tuned checkpoint:
\begin{align}
 R_s(P)
 &=\frac{\operatorname{PPL}(\thetaun^{(s)};P)}
 {\operatorname{PPL}(\thetaft^{(s)};P)}
 =\exp\!\left(\Lloss(\thetaun^{(s)};P)
 -\Lloss(\thetaft^{(s)};P)\right),\nonumber\\
 \mathsf{Util}_s&=\mathbf 1\!\left\{\max_P R_s(P)\le1.10\right\}.
 \label{eq:utility-predicate}
\end{align}
Here $R_s(P)=1$ means unchanged perplexity. The utility criterion permits at
most a $10\%$ increase on each population; an improvement, $R_s(P)<1$, also
passes.

\subsection{Run-validity guard and checkpoint verdict}

Define $\mathsf{RunOK}_s=1$ when the run completes without a nonfinite loss
or a batch loss above 20, and $0$ otherwise. A criterion with a missing or
nonfinite input is false. On an admitted membership axis, the seed-level
checkpoint verdict is
\begin{equation}
 \mathsf{Pass}_s=
 \mathsf{Mem}_s\land\mathsf{Sel}_s\land
 \mathsf{Util}_s\land\mathsf{RunOK}_s.
 \label{eq:joint-verdict}
\end{equation}
We evaluate each seed independently so that strong runs cannot compensate for
a failed checkpoint. We report the number of seeds passing each criterion and
the conjunction. A setting passes if every seed passes, fails if every seed
fails, and is inconclusive otherwise. Across-seed means remain descriptive:
we never apply the verdict to an averaged checkpoint that does not exist.

\subsection{Detector definitions and access}

The four operational detectors map to the access bands in
Section~\ref{sec:setting}; each stronger band includes the observations of the
preceding bands.
\begin{itemize}
 \setlength{\itemsep}{0pt}
 \setlength{\parsep}{0pt}
 \setlength{\topsep}{2pt}
 \item \emph{Released-model loss}, $\Lloss(\thetaun;z)$, uses only
 $\mathcal O_{\mathrm{rel}}$.
 \item \emph{Two-checkpoint loss change},
 $\Lloss(\thetaun;z)-\Lloss(\thetaft;z)$, uses
 $\mathcal O_{\mathrm{2cp}}$ and isolates movement across the update.
 \item \emph{Directional alignment} uses $\mathcal O_{\mathrm{wb}}$ and is the
 signed cosine between $\nabla_\theta\Lloss(\thetaun;z)$ and
 $\thetaun-\thetaft$.
 \item \emph{Calibrated gradient magnitude} also uses
 $\mathcal O_{\mathrm{wb}}$ and is the ratio of squared gradient norms at
 $\thetaun$ and $\thetaft$, with the denominator floored by $\epsilon$.
\end{itemize}
We compare detectors only under matched access; performance gains obtained
from an additional checkpoint or gradients are not attributed to a better
score alone.

\subsection{Low-prevalence operating point}

The checkpoint verdict and detector usefulness answer different questions.
When deleted records are rare, even a small false-positive rate can make many
flags incorrect. If the score distributions remain fixed, changing only class
prevalence leaves ROC-AUC unchanged but changes precision. Let $\pi$ be the
deleted-record fraction, $t$ the true-positive rate, and $f$ the
false-positive rate. Then
\begin{equation}
 \mathrm{Prec}(\pi)=\frac{\pi t}{\pi t+(1-\pi)f}.
 \label{eq:prevalence}
\end{equation}
For each seed, we flag the $k$ highest-scoring evaluation records, where $k$
is the known number of deleted records. We insert the resulting $t$ and $f$
into Eq.~\eqref{eq:prevalence} and project precision to one deleted record per
100 non-deleted records ($\pi=1/101$). This is not a deployment measurement:
it assumes that the measured $t$ and $f$ remain unchanged at the projected
prevalence. If no false positives occur among the $n_{-}$ evaluated
non-deleted records, we substitute $f=1/n_{-}$ rather than infer perfect
precision from a finite sample. This conservative substitution is a reporting
convention, not a confidence bound.

\section{Experimental Setup}
\label{sec:experiments}

\subsection{Corpora, models, and populations}

We study ArxivMIA and BookMIA with Llama-3.2-1B and Qwen2.5-1.5B
\cite{zhliu2024arxivmia,shi2024detecting}. Exact retraining covers all four
corpus--model pairings; approximate methods are audited on ArxivMIA/Llama and
BookMIA/Qwen. Per seed, 1,000 benchmark label-1 records form 200 forget and
800 retain records; 600--800 retain records are scored. The held-out utility
set $D_{\mathrm{test}}$ contains 1,000 sampled label-0 records, 300 of which
are scored and serve as the original membership comparison. Scored test-token
counts have mean (10th--90th percentile) 214 (96--340) on ArxivMIA/Llama and
511 (511--511) on BookMIA/Qwen. The post-hoc BookMIA diagnostic instead uses
300 disjoint, never-fine-tuned label-1 records.

All scored BookMIA sequences reach the 512-token cap, versus only two to five
of 1,300 ArxivMIA sequences per seed. Corpus and truncation therefore co-vary;
we do not infer an isolated causal corpus effect.

Both models use QLoRA (NF4, rank 16, scaling 32, dropout 0.05) with
six epochs and effective batch 32. Exact retraining restricts the same
schedule to the 800 retain records; logged fine-tuning and retraining runs
take 186 and 150 optimizer steps, respectively. Full settings are in the
artifact.

\subsection{Approximate methods}

We evaluate three approximate-unlearning families. \textbf{GA+GD} alternates
gradient ascent on $\Dforget$ with gradient descent on $\Dretain$.
\textbf{NPO-RT} combines a bounded NPO forget loss against a frozen
$\thetaft$ reference with retain training~\cite{zhang2024negative}.
\textbf{SCRUB} trains a student to match the $\thetaft$ teacher on retain
records while diverging on forget records~\cite{kurmanji2023towards}. Each
method runs for at most 100 updates. ArxivMIA/Llama GA+GD and SCRUB runs
are confirmatory, and NPO-RT is preregistered. BookMIA/Qwen SCRUB is
exploratory; the narrow GA+GD refinement is post-hoc. Full grids and
configuration counts are in the artifact.

\subsection{Evaluation protocol}

We evaluate three training seeds and assign each criterion and the conjunctive
verdict to each checkpoint independently. Across-seed means and sample
standard deviations are descriptive summaries and are not combined to
determine a verdict. Any sample bootstrap quantifies within-checkpoint rather
than across-training uncertainty. The audit criteria and thresholds were fixed before confirmatory runs and
reused unchanged across datasets. The matched-population diagnostic is
post-hoc. The artifact contains scores, configurations, and analysis code.

\section{Results}
\label{sec:results}

\subsection{Reference Calibration Exposes a Population Mismatch}
\label{sec:reference-results}

\begin{figure*}[t]
\centering
\includegraphics[width=0.95\textwidth]{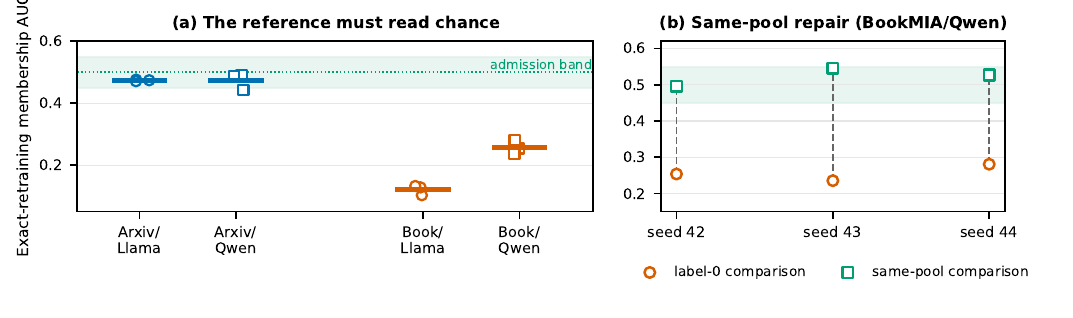}
\caption{\textbf{The comparison population determines reference validity.}
(a) Exact-retraining AUC across four corpus--model cells; points are seeds,
bars are means, and shading marks the admission band. (b) Post-hoc
BookMIA/Qwen diagnostic with the same forget records, checkpoints, and fixed
gate, but a same-pool, never-fine-tuned comparison population.}
\label{fig:reference}
\end{figure*}

The reference gate admits a membership comparison only when exact-retraining
AUC lies within the $[0.45,0.55]$ chance region. Figure~\ref{fig:reference}a
shows that ArxivMIA passes: AUC is $0.474\pm0.003$ for Llama and
$0.473\pm0.027$ for Qwen. BookMIA does not: AUC is $0.122\pm0.016$ and
$0.257\pm0.023$, respectively. Because exact retraining removes the forget
records by construction, we invalidate the original BookMIA membership axis
rather than call this a deletion failure.

The pattern occurs for both model families, whereas the two ArxivMIA cells
are nearly identical. It therefore points away from a model-specific cause,
but does not establish a causal corpus effect because corpus, comparison
population, and truncation vary together.

\textbf{A matched comparison isolates the mismatch.} BookMIA's benchmark
label-1 and label-0 records come from different book populations and likely
encode base-model pretraining membership. Our target property is instead
membership in the added fine-tuning set. The original comparison thus varies
both fine-tuning membership and benchmark population.

Holding the checkpoints and threshold fixed, we replace label-0 records with
unused label-1 records from the same source pool. BookMIA/Qwen
exact-retraining AUC then moves from $0.257\pm0.023$ to $0.523\pm0.025$
(Figure~\ref{fig:reference}b). This supports comparison-population mismatch
in the tested cell: the original score separates benchmark populations as
well as fine-tuning membership. The diagnostic is post-hoc and was not run
on BookMIA/Llama, so it does not establish that all BookMIA membership metrics
fail.

The base-model preflight does not expose this problem: before fine-tuning,
member-versus-label-0 AUC is approximately $0.460$ on BookMIA and $0.463$ on
ArxivMIA. The mismatch becomes consequential after fitting on $\Dretain$.
Thus, a benign base-only check cannot replace calibration on the
exact-retraining reference.

\textbf{Reference validity alone is insufficient.} Against the matched
BookMIA holdout, exact retraining reads $0.523$ and aggressive GA+GD reads
$0.501$. Yet GA+GD collapses on two seeds and violates utility on all three.
Membership alone therefore cannot separate deletion from global damage; the
next stage adds targeted-shift and utility checks.

\subsection{No Tested Approximate Configuration Passes}
\label{sec:audit-results}

\textbf{Controls anchor the audit scale.} Table~\ref{tab:controls} reports
forget-set TSF for four controls. Exact retraining equals one by construction,
whereas the null refit remains near zero on both corpora; training randomness
is therefore not the dominant source of the exact-retraining denominator in
these runs. A routine update moves TSF despite deleting nothing, and the same
update after exact retraining shifts TSF below one. Routine maintenance can
therefore move the empirical deletion reference.

\begin{table}[t]
\centering
\caption{Control calibration: mean forget-set TSF $\pm$ across-seed SD.}
\label{tab:controls}
\footnotesize
\begin{tabular}{@{}lccc@{}}
\toprule
Condition & Deletes? & ArxivMIA & BookMIA \\
\midrule
Null refit & no & $0.001\pm0.014$ & $-0.008\pm0.018$ \\
Routine update & no & $0.132\pm0.009$ & $0.063\pm0.014$ \\
Exact retrain & yes & $1.000\pm0.000$ & $1.000\pm0.000$ \\
Exact + update & yes & $0.911\pm0.007$ & $0.947\pm0.006$ \\
\bottomrule
\end{tabular}
\end{table}

\begin{figure*}[t]
\centering
\includegraphics[width=0.95\textwidth]{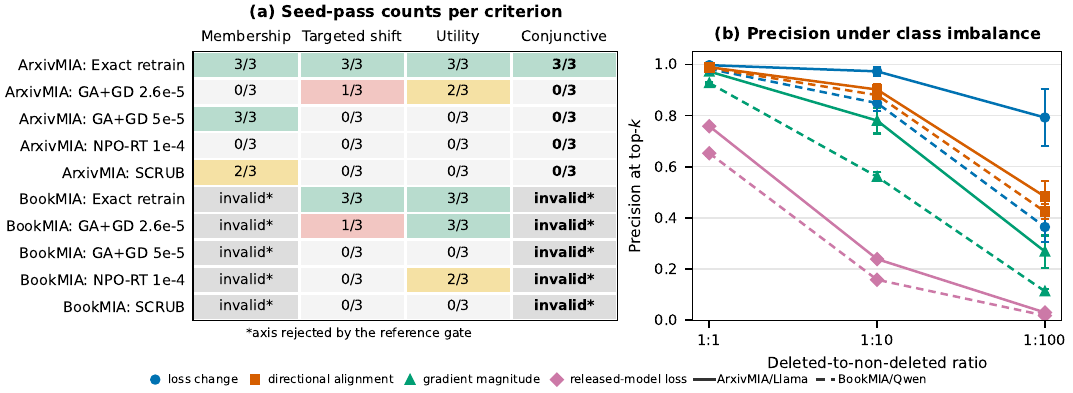}
\caption{\textbf{Reference-gated method audit.} (a) Per-criterion seed-pass
counts; conjunction requires one checkpoint to pass all criteria. Original
BookMIA membership and conjunction are invalid because the reference gate
fails. (b) Top-$k$ precision by deletion prevalence; bars show across-seed SD.}
\label{fig:audit}
\end{figure*}

\textbf{The conjunction changes the method verdict.}
Figure~\ref{fig:audit}a reports marginal pass counts and the number of seeds
that pass every criterion on the same checkpoint. Exact retraining passes all
criteria for all three ArxivMIA/Llama seeds. On BookMIA/Qwen it passes
targeted shift and utility, but the original membership axis is invalid
(Section~\ref{sec:reference-results}); this is not counted as a deletion
failure.

None of the tested GA+GD, NPO-RT, or SCRUB configurations passes the
conjunction on the valid ArxivMIA/Llama axis; the artifact reports the full
grid. The invalid original BookMIA/Qwen axis cannot support a joint method
verdict. Even under its matched-population diagnostic, aggressive GA+GD
reaches membership AUC $0.501$ but fails targeted shift and utility. Across
the evaluated settings, weaker GA+GD moves too little, stronger GA+GD damages
utility, and NPO-RT and SCRUB never satisfy all criteria together on the valid
axis. This result is limited to the tested configurations, not the method
families in general.

\textbf{Single-axis and averaged results can mislead.} In the GA+GD strength
sweep, four of seven ArxivMIA settings from $3.5\times10^{-5}$ through
$5\times10^{-5}$
fall inside the membership chance region by across-seed mean. All four fail
utility, with mean retained-perplexity ratios from $12.74$ to
$4.93\times10^7$. Membership alone would therefore label model damage as
successful forgetting.

The post-hoc $2.6\times10^{-5}$ refinement exposes a second aggregation
hazard: its across-seed means satisfy the targeted-shift and utility
thresholds, but only one checkpoint satisfies both and none passes
membership. Marginal means cannot replace a within-checkpoint conjunction.

\subsection{Operational Capability and Cost}
\label{sec:operational}

The preceding audit evaluates checkpoints; we now evaluate the detectors
used to identify individual deleted records. We ask how they behave when
deletions are rare and what access and computation they require. To isolate
detector capability from approximate-unlearning failure, both analyses use
exact retraining and distinguish deleted from retained training records.

\textbf{High AUC does not ensure precision for rare deletions.}
ROC-AUC measures ranking across thresholds but ignores deletion prevalence.
Figure~\ref{fig:audit}b fixes each seed's measured top-$k$ true- and
false-positive rates and varies only the assumed prevalence using
Eq.~\eqref{eq:prevalence}. Precision falls for every detector as deletions
become rarer. For the two leading detectors, ROC-AUC remains between $0.9936$
and $0.9999$, yet projected precision at $1{:}100$ is only $0.365$--$0.793$.
Thus, near-perfect ranking does not imply that a flagged record is likely to
be deleted. These values are projections, not deployment measurements.

\begin{figure*}[t]
\centering
\includegraphics[width=0.90\textwidth]{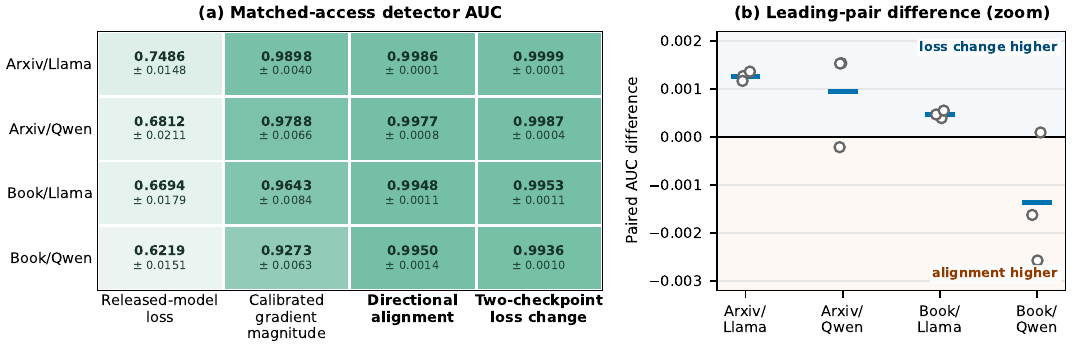}
\caption{\textbf{Matched-access detector comparison.} (a) Mean
deletion-history ROC-AUC $\pm$ across-seed SD over three exact-retraining runs;
rows are corpus--model cells, columns are detectors, and darker shading
indicates higher mean AUC. (b) The paired within-seed AUC difference between
the leading detectors is small and changes sign; circles are seeds and bars
are means.}
\label{fig:operational}
\end{figure*}

\textbf{Without a consistent performance advantage, operational cost guides
the choice.} Two-checkpoint loss change and directional alignment are the two
leading detectors in every corpus--model cell (Figure~\ref{fig:operational}a),
while calibrated gradient magnitude and released-model loss rank lower.
However, the absolute difference between the two leaders' mean AUCs is at
most $0.0014$, and its direction changes across cells and seeds
(Figure~\ref{fig:operational}b). With three seeds per cell, these results
identify a leading pair but establish neither a consistent winner nor
statistical equivalence. Detection performance therefore provides no stable
basis for choosing between them; their access and computational requirements
become the practical differentiators.

\begin{table}[t]
\centering
\caption{Detector requirements and measured standalone evaluation time. Time
is in seconds per scored sample, composed from median primitive measurements
over 32 samples and five repeats on one RTX A5000; checkpoint loading is
excluded.}
\label{tab:cost}
\footnotesize
\begin{tabular}{@{}p{0.42\columnwidth}rrr@{}}
\toprule
Detector & \shortstack{Check-\\points} & \shortstack{Backward\\passes} &
\shortstack{Time\\(s/sample)} \\
\midrule
Released-model loss & 1 & 0 & $0.033$ \\
Two-checkpoint loss change & 2 & 0 & $0.067$ \\
Directional alignment & 2 & 1 & $0.108$ \\
Calibrated gradient magnitude & 2 & 2 & $0.220$ \\
\bottomrule
\end{tabular}
\end{table}

Table~\ref{tab:cost} makes this operational difference explicit. Both leading
detectors require the pre- and post-unlearning checkpoints, but loss change
requires no per-example backward pass and takes $0.067$ s/sample on our
hardware; directional alignment requires one backward pass and takes $0.108$
s/sample. Calibrated gradient magnitude is more expensive still at $0.220$
s/sample. Thus, when both checkpoints are available, we favor loss change as
the lower-cost member of the leading pair---not as a statistically superior
detector. These timings are hardware-specific engineering measurements, not
complexity bounds. This operational choice also does not replace reference
calibration: Table~\ref{tab:controls} shows that routine maintenance can move
TSF without deleting records.

\section{Discussion and Limitations}
\label{sec:discussion}

\textbf{Validate the measurement before judging the method.}
A deletion verdict is meaningful only if its comparison behaves correctly on
a known-ground-truth reference. Exact retraining provides this test because
requested records are absent by construction. If the reference gate fails,
the axis cannot judge exact or approximate deletion. Post-hoc repairs may
diagnose the mismatch, but remain setting-specific and non-confirmatory; they
do not invalidate the benchmark or membership inference generally.

\textbf{Membership and deletion history answer different questions.}
Current membership asks whether the released model separates forget records
from a valid never-fine-tuned population. Chance performance says only that
this detector cannot separate them; it does not rule out stronger membership,
reconstruction, or extraction attacks. Deletion-history detection instead
asks which known members were targeted, often using pre- and post-unlearning
checkpoints. This access can support an authorized audit, but may expose
sensitive history to an insider who also obtains a candidate
pool~\cite{deepak2025fuma}. We evaluate authorized capability, not adversarial
acquisition of these inputs.

\textbf{Method conclusions remain deliberately scoped.}
The verdict is confined to the evaluated corpora and grids, two 1--1.5B model
families, and rank-16 QLoRA. It establishes neither a general failure of
GA+GD, NPO-RT, or SCRUB nor transfer to larger or full-parameter models.
Because corpus, comparison population, and truncation co-vary, we do not infer
a causal corpus effect. Future comparisons should validate exact retraining
before applying every criterion to the same checkpoint.

\textbf{Statistical and operational limits remain.}
Three independent runs support checkpoint-level verdicts, but not statistical
equivalence or population-wide detector rankings. Fixed engineering
tolerances and one calibrated loss-ratio score do not replace complementary or
adaptive membership attacks~\cite{carlini2022membership,
naderloui2025rectifying}. Exact retraining uses fewer optimizer steps on the
smaller retain set, so the reference comparison does not isolate deletion
from optimization budget. The 1:100 analysis projects
prevalence rather than estimating deployment performance, and single-GPU
timings support only relative cost claims. Priorities for broader validation
are larger and full-parameter models, factorial population/truncation tests,
adaptive attacks, step-matched retraining, and realistic prevalence and access
conditions.

\section{Conclusion}
\label{sec:conclusion}

A fine-tuning-unlearning verdict is only as credible as the counterfactual
that gives its metric meaning. Exact retraining exposed a misaligned
membership comparison; null refits and routine updates measured
non-deletion-induced loss change; and within-seed conjunctive criteria
prevented model damage from being mistaken for targeted forgetting. Under the
reference-gated audit, none of the tested approximate configurations
satisfied membership removal, targeted shift, and utility preservation
together. The central lesson: a score near chance is not evidence of
forgetting until the reference establishes what chance means.

\section*{Acknowledgment}
Mayank Gulati and Gerhard Wunder were supported by the German Federal Ministry of Education
and Research (BMBF), programme ``Souver\"an. Digital. Vernetzt.'', through
AIgenCY (16KIS2013), 6G-RIC (16KISK025 and 16KISK020K), and UltraSec
(16KIS1682). Gerhard Wunder was also supported by the German Research Foundation
(DFG), SPP 2378 ``ResNets: Resilience in Connected Worlds'' (WU 598/12-1).

\bibliographystyle{IEEEtran}
\bibliography{references}

\end{document}